%% file: main.tex
\documentclass[10pt,twocolumn,letterpaper]{article}

\usepackage[pagenumbers]{cvpr} 

\input{preamble}

\definecolor{cvprblue}{rgb}{0.21,0.49,0.74}
\usepackage[pagebackref,breaklinks,colorlinks,allcolors=cvprblue]{hyperref}

\usepackage[most]{tcolorbox}
\tcbset{
    theoremstyle/.style={
        enhanced,
        colback=gray!10,
        colframe=gray!40,
        boxrule=0.3pt,
        arc=2pt,
        left=6pt,
        right=6pt,
        top=6pt,
        bottom=6pt
    }
}

\def\paperID{8270} 
\def\confName{CVPR}
\def\confYear{2026}

\title{FREE: Uncertainty-Aware Autoregression for Parallel Diffusion Transformers}

\author{
Xinwan Wen\\
Tsinghua University\\
{\tt\small wenxw24@mails.tsinghua.edu.cn}
\and
Bowen Li\\
Tsinghua University\\
{\tt\small libw25@mails.tsinghua.edu.cn}
\and
Jiajun Luo\\
Tsinghua University\\
{\tt\small luo-jj24@mails.tsinghua.edu.cn}
\and
Ye Li\\
Tsinghua University\\
{\tt\small liye23@mails.tsinghua.edu.cn}
\and
Zhi Wang\\
Tsinghua University\\
{\tt\small wangzhi@sz.tsinghua.edu.cn}
\thanks{Corresponding author.}}

\begin{document}
\maketitle
\input{sec/0_abstract}    
\input{sec/1_intro}
\input{sec/2_related_work}
\input{sec/3_method}
\input{sec/4_experiments}
\input{sec/5_conclusion}
{
    \small
    \bibliographystyle{ieeenat_fullname}
    \bibliography{main}
}

\input{sec/X_suppl}

\end{document}

%% file: preamble.tex
\usepackage{amsmath, amssymb}
\usepackage[ruled,vlined,linesnumbered]{algorithm2e}
\usepackage{xcolor}

\usepackage[numbers]{natbib}

%% file: sec/0_abstract.tex
\begin{abstract}
Diffusion Transformers (DiTs) achieve state-of-the-art generation quality but require long sequential denoising trajectories, leading to high inference latency.
Recent speculative inference methods enable lossless parallel sampling in U-Net–based diffusion models via a drafter–verifier scheme, but their acceleration is limited on DiTs due to insufficient draft accuracy during verification.
To address this limitation, we analyze the DiTs' feature dynamics and find the features of the final transformer layer (top-block) exhibit strong temporal consistency and rich semantic abstraction. Based on this insight, we propose \textbf{FREE}, a novel framework that employs a lightweight drafter to perform feature-level autoregression with parallel verification, guaranteeing lossless acceleration with theoretical and empirical support.
Meanwhile, prediction variance (uncertainty) of DiTs naturally increases in later denoising steps, reducing acceptance rates under speculative sampling. To mitigate this effect, we further introduce an uncertainty-guided relaxation strategy, forming \textbf{FREE (relax)}, which dynamically adjusts the acceptance probability in response to uncertainty levels.
Experiments on ImageNet-$512^2$ show that \textbf{FREE} achieves up to $1.86\times$ acceleration, and \textbf{FREE (relax)} further reaches $2.25\times$ speedup while maintaining high perceptual and quantitative fidelity in generation quality.
\end{abstract}

%% file: sec/1_intro.tex
\section{Introduction}
\label{sec:intro}

\begin{figure*}[t]
  \centering
  \includegraphics[width=\textwidth]{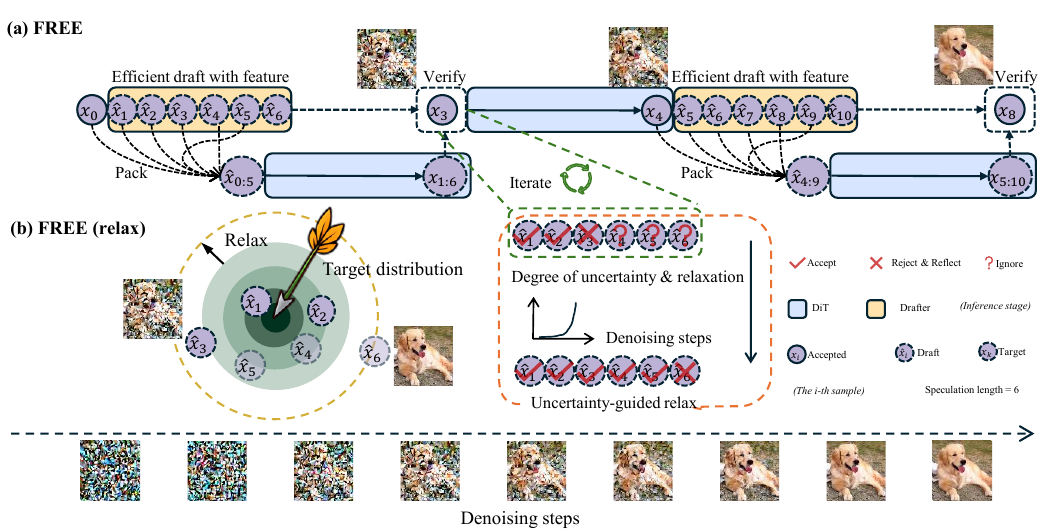}
  \caption{
    Overview of the \textbf{FREE} framework.
    \textbf{(a)} In \textbf{FREE}, a lightweight draft model autoregressively generates six speculative drafts in the top-block feature space. The first five drafts, together with the input, are packed and sent to the target DiT for one pass to obtain the reference outputs. Each draft is sequentially verified against the reference: accepted drafts are kept; Upon rejection, a reflection coupling is triggered and all subsequent drafts are ignored.
    \textbf{(b)} In \textbf{FREE (relax)}, an uncertainty-aware relaxation mechanism is introduced to dynamically adjust the acceptance probability, avoiding over-constraining the draft model under high uncertainty.
  }
  \label{fig:framework}
\end{figure*}

Diffusion probabilistic models~\cite{DDPM, latent, song2021scorebased} have demonstrated remarkable generation quality and broad applicability, providing a unified framework for high-fidelity visual and multimodal synthesis~\cite{diff23D, policy, protein}.
These models generate samples through multi-step denoising, progressively refining the latent toward the data manifold.
While many acceleration techniques~\cite{DDIM, distill} have effectively reduced the number of denoising iterations, the process remains sequential, limiting scalability and preventing parallel execution.

Recently, Speculative inference frameworks~\cite{sd_origin, cai2024medusa, reflection, ASD} have emerged as a promising paradigm to break this sequential bottleneck, enabling lossless and parallel acceleration.
These methods employ a drafter–verifier scheme, where a lightweight drafter proposes multiple future denoising steps that are verified in parallel by a target model.
This approach achieves substantial acceleration while preserving fidelity, and has shown strong performance on pixel-space diffusion models such as Denoising Diffusion Probabilistic Models (DDPMs)~\cite{DDPM, moradi2025rddpm, teng2025fingerprinting} and Latent Diffusion Models (LDMs)~\cite{latent, zhang2025diffusion, jeong2025latent}.
However, existing speculative inference techniques have yet to generalize to DiTs~\cite{DiT, mo2023dit, zhang2025magicmirror}, where global re-contextualization at each denoising step leads to abrupt, non-smooth feature shifts across timesteps, unlike the gradual and localized feature evolution in U-Net architectures~\cite{u-net} used in prior work. Under these dynamics, even small prediction errors are rapidly amplified, sharply reducing draft acceptance rates and rendering existing speculative strategies ineffective when applied directly to DiTs (see~\cref{tab:parallel_efficiency}).
This raises a central question: \emph{How can we make speculative sampling work effectively for DiTs?}

To answer this question, we first analyze the feature dynamics of DiTs (see~\cref{fig:top_block}) and identify two key characteristics:
(1) Temporal Stability of Intermediate Features.
Intermediate hidden representations are more temporally stable than final prediction outputs, showing higher temporal cosine similarity across adjacent timesteps.
(2) Semantic Richness of Top-Block Features.
Higher transformer blocks exhibit stronger semantic transformation, with larger activation changes between their input and output features.
These two properties indicate that top-block features are both temporally stable and semantically rich, making them ideal for feature-level autoregressive multi-step drafting.
We further analyze the prediction behavior of DiTs and derive a third property:
(3) Uncertainty Growth Along the Denoising Trajectory.
Prediction deviation increases at later timesteps, leading to progressively higher uncertainty and reduced acceptance rates under speculative sampling (see~\cref{fig:acc_cha}).

Building on these insights, we propose FREE, a speculative inference framework that pairs a lightweight drafter and the target DiT. The drafter autoregressively predicts future top-block features, while the target model verifies multiple denoising steps in parallel, achieving substantial lossless acceleration with both theoretical and empirical support.
To address the increasing uncertainty of DiTs in later stages, we further introduce an uncertainty-guided relaxation strategy, forming FREE (relax), which dynamically adjusts the acceptance probability of drafts, achieving faster inference without perceptible degradation.

We evaluate FREE on ImageNet-$256^2$ and $512^2$ benchmarks~\cite{russakovsky2015imagenet}, achieving up to $1.82\times$ and $1.86\times$ wall-clock speedups over standard DiT sampling. 
The relaxed variant, FREE (relax), further improves acceleration to $2.19\times$ and $2.25\times$ while maintaining comparable perceptual quality across both resolutions.

%% file: sec/2_related_work.tex
\section{Related Work}
\label{sec:related_work}

\subsection{Diffusion Transformers (DiTs)}
DiTs~\cite{DiT, wang2025lavin, feng2025dit4edit, dasari2025ingredients} replace convolutional backbones with global self-attention, enabling strong semantic aggregation in latent representations. This structure improves generative fidelity and scalability but introduces high inference latency from deep attention computation.
Recent studies explore inference acceleration for DiTs~\cite{ma2024learning, chen2025accelerating, chu2025omnicache, zhang2025blockdance}.
Efficient attention mechanisms~\cite{att1, att2, dao2024flashattention, yuan2024ditfastattn} reduce the quadratic complexity of self-attention, and token pruning~\cite{prun1, liang2022evit, prun3, fang2025tinyfusion} skip or remove redundant tokens to improve throughput.
These architectural methods retain the original diffusion trajectory but reduce FLOPs per step.
A complementary line of research accelerates sampling by reducing denoising steps. Deterministic or ODE-based samplers~\cite{DDIM, dpm+, unipc, ode5steps} reformulate the diffusion process into deterministic trajectories, while distillation-based approaches~\cite{distill, onestep, momentMatch, datafree} compress long diffusion trajectories into compact student models for few-step generation.

Despite efficiency, these approaches still inherit the strictly sequential structure of the diffusion process, where each denoising step depends on the output of the previous one. 
Consequently, inference speed scales linearly with the number of steps, fundamentally limiting parallelization and preventing full utilization of modern massively parallel hardware.
Classical analyses of score-based sampling also indicate that approximately $\tilde{O}(d)$ iterations are required to sample from $d$-dimensional distributions~\cite{stepsN}, highlighting the difficulty of achieving truly few-step generation without altering the generative objective or model structure.
Thus, further acceleration requires breaking the stepwise dependency, rather than simply reducing per-step cost or trajectory length. Our approach follows this orthogonal direction.

\subsection{Speculative Inference in Diffusion Models}
Speculative inference, inspired by large language models~\cite{sd_origin, cai2024medusa, li2024eagle, LLM_spec, xia-etal-2024-unlocking}, accelerates autoregressive generation by producing lightweight drafts and verifying them in parallel with the target model.
Building on this paradigm, \textit{Accelerated Diffusion Models via Speculative Sampling}~\cite{reflection} and \textit{AutoSpeculation}~\cite{ASD} extend the idea from discrete token generation to the continuous denoising process, employing reflection-based maximal coupling~\cite{coupling} to preserve the target sampling distribution.

These methods differ primarily in how drafts are generated.
The former employs either a lightweight independent drafter or frozen-target drafting, where multiple future steps are approximated by reusing a single score evaluation from the target model.
The latter performs auto-speculative multi-step rollout by leveraging the hidden exchangeability property of diffusion trajectories to parameterize their proposal distributions.
While effective in U-Net–based diffusion models, both strategies exhibit pronounced acceptance degradation when applied to DiTs: independent/frozen drafting produces coarsely approximated proposals, whereas the exchangeability assumption required by auto-speculative rollout does not hold strongly in DiT architectures, leading to unstable multi-step proposals.
This motivates drafting in a more stable and semantically coherent feature space, better aligned with DiT’s internal representation dynamics.
A recent feature-space speculative approach, SpeCa~\cite{SpeCa}, takes this direction by caching hidden features and using a Taylor expansion for drafting, with proposals accepted based on a feature-discrepancy threshold. While effective, it is hard to roll back once denoising errors accumulate and relies on approximate verification, introducing a tunable trade-off between speed and sampling fidelity.
This suggests that feature space is a promising domain for speculative decoding, but achieving lossless sampling requires a principled acceptance mechanism that avoids heuristic thresholds.

%% file: sec/3_method.tex
\section{Method}
\label{sec:method}

\begin{figure}[t]
  \centering
   \includegraphics[width=\linewidth]{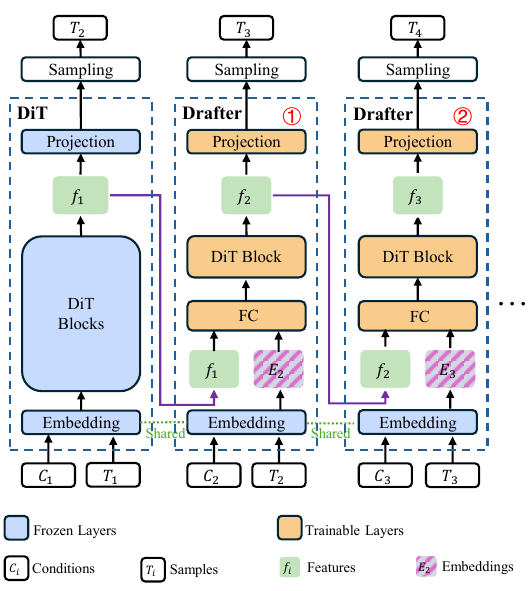}

   \caption{
   Architecture of the proposed FREE framework.
   \textbf{Left:} A single forward pass of DiT. \textbf{Right:} The feature-level autoregressive process of the drafter.
    \textcolor{red}{Red} numbers indicate the index of each draft forward pass.
    ``Sampling'' denotes drawing samples from the predicted distribution based on the model outputs.
   }
   \label{fig:architecture}
\end{figure}

In this section, we detail the FREE framework (~\cref{fig:framework}), which frees DiT sampling from strict sequential execution via feature-level speculative inference.
We first outline how multi-step drafting is performed directly in the top-block feature space, forming a compact autoregressive pathway. 
We then describe how the drafter is trained to follow the feature evolution of the frozen DiT for reliable future-step proposals.
Next, we show how the drafts are verified in parallel via reflection-based coupling, ensuring exact consistency with the target model.
Finally, we present an uncertainty-aware relaxation mechanism that adjusts verification strictness according to timestep-dependent prediction variance, improving acceptance rates without perceptual degradation.

\begin{algorithm}[t]
\caption{FREE}
\label{alg:free}
\SetAlgoLined
\KwIn{Total denoising steps $T$, speculation length $L$, target model $q$, draft model $p$, and variance schedule $\sigma^2_{1:T+1}$}

Sample $x_0 \sim \mathcal{N}(x; 0, I_d)$ and set $a \leftarrow 1$ \\

\While{$a \le T$}{
    Compute mean $m_a$ and feature $f_a$ of $q_a = q(\cdot \mid x_{a-1})$ \\
    Sample $x_a \sim \mathcal{N}(x; m_a, \sigma^2_a I_d)$ \\
    Set $b \leftarrow \min(T, a+L)$, $\hat{x}_a \leftarrow x_a$ \\

    \For{$t = a+1 : b+1$}{
        Compute mean $\hat{m}_t$ and feature $f_t$ of $p_t = p(\cdot \mid \hat{x}_{t-1}, f_{t-1})$ \\
        Sample $\hat{x}_t \sim \mathcal{N}(x; \hat{m}_t, \sigma^2_t I_d)$ \\
    }

    Compute means $m_{a+1:b+1}$ of $q(\cdot \mid \hat{x}_{a:b})$ \textbf{in parallel} \\

    \For{$k = a+1 : b+1$}{
        $(x_k, \text{accepted}) \leftarrow \textsc{Verification}(\hat{m}_k, m_k, \sigma^2_k, \hat{x}_k)$ \\
        \If{not(accepted )}{
            break \\
        }
    }
    $a \leftarrow k$ \\
}
\Return{$x_{0:T+1}$} \\
\end{algorithm}

\begin{figure}[t]
  \centering
  \includegraphics[width=\columnwidth]{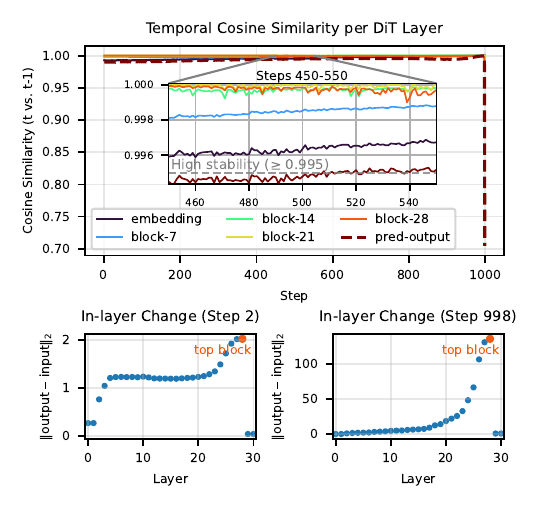}
  \caption{
  Temporal stability and in-layer feature change across DiT layers during diffusion sampling.
  \textbf{Top}: Temporal cosine similarity between features at consecutive steps ($t$).
  Transformer blocks maintain high similarity (\(\ge 0.995\)) throughout most denoising steps, especially the top block (\textcolor{orange}{block-28}), while the final output head (\textcolor{red!60!black}{pred-output}) shows lower temporal stability.
  \textbf{Bottom}: In-layer feature change $\|\mathrm{output} - \mathrm{input}\|_2$ at early (step 2) and late (step 998) denoising stages. The top block (\textcolor{orange}{layer 28}) exhibits the largest change at both stages, consistent with it being the most semantically expressive layer.
  }
  \label{fig:top_block}
\end{figure}

\subsection{Feature-Level Autoregressive Drafting}
\label{drafting}
As illustrated in~\cref{fig:architecture}, FREE consists of a target DiT and a lightweight drafter.
At each denoising step, the drafter receives the current latent sample together with its top-block feature and predicts the next-step feature and noise estimate, the latter of which is then converted into the mean of the reverse diffusion transition~\cite{DDPM}. This allows FREE to forecast multiple future denoising steps before the target model is invoked again, reducing costly DiT evaluations.

To detail the inference procedure (see~\cref{alg:free}), the target model first performs a standard forward pass, producing the feature $f_a$ and reverse mean $m_a$, from which the reference latent $x_a$ is drawn according to a Gaussian distribution $\mathcal{N}(m_a, \sigma_a^2 I_d)$.
Starting from this accurate reference state, the drafter then autoregressively predicts the next $L$ denoising iterations, where $L$ is a user-chosen speculation length.
Let $b = \min(T, a+L)$, with $T$ the total number of denoising steps.
For each step $t$ from $a+1$ to $b$, the drafter predicts the feature $f_t$ and mean $\hat{m}_t$ based on $(\hat{x}_{t-1}, f_{t-1})$ from the previous timestep, and then draws the sample $\hat{x}_t$ from $\mathcal{N}(\hat{m}_t, \sigma_t^2 I_d)$.
This produces a sequence of $L$ speculative latent states and feature trajectories:
\begin{equation}
  \{(\hat{x}_{a+1}, f_{a+1}), \cdots , (\hat{x}_b, f_b)\},
  \label{eq:drafts}
\end{equation}
which is subsequently verified in parallel.

A central design choice in FREE is the use of the top-block feature as the autoregressive state.
As shown in~\cref{fig:top_block}, internal features exhibit smoother temporal evolution compared to the final prediction output, which represents independent Gaussian noise at each step and therefore shows low temporal similarity.
Among them, the top-block feature maintains consistently high temporal correlation across most denoising steps, reflecting a stable and predictable evolution pattern.
At the same time, it undergoes comparatively larger in-layer transformations at both early and late stages, indicating that it remains semantically active rather than merely propagating features forward.
These complementary properties---temporal stability and semantic richness---make it well suited for feature-level autoregression, enabling the drafter to reliably forecast future latent states while avoiding the instability of noisy prediction outputs and the limited semantics of low-level features.

To maintain practical efficiency, the drafter is implemented as a lightweight surrogate of the target model (\eg DiT-XL/2 with 28 transformer blocks), reusing the shared embedding layer and producing the draft noise prediction through a small output projection.
Between them, a fully connected layer (FC layer) fuses the embedding output with the propagated features, and the resulting fused representation is then processed by a single DiT block.
The computational cost of each DiT block is dominated by the self-attention operation, which scales quadratically with the token sequence length, whereas the FC layer has linear complexity.
As the sequence length increases, attention dominates in both models, and the relative computational overhead of the drafter asymptotically approaches $\sim 1/28$ of the target DiT, accounting for only a small fraction of the total sampling cost.
This design achieves substantial computational savings while remaining fully compatible with the original DiT architecture, enabling efficient speculative inference within the FREE framework.

\begin{algorithm}[t]
\caption{\textsc{Verification} (adapted from~\cite{reflection})}
\label{alg:verification}
\KwIn{Proposal mean $\hat{m}$, target mean $m$, variance $\sigma^2$, and proposal sample $\hat{x}$.}

Sample $u \sim \mathrm{Uniform}(0,1)$ \\
Set $\text{bool} \leftarrow \mathbb{I}\!\left[
    u \le \min\!\left(1,
    \mathcal{N}(\hat{x};\,m,\,\sigma^2 I_d)/\mathcal{N}(\hat{x};\,\hat{m},\,\sigma^2 I_d)
    \right)
\right]$ \\

\If{$\text{bool}$}{
    $x \leftarrow \hat{x}$ \\
}
\Else{
    Set $e = (\hat{m} - m)/(\|\hat{m} - m\|)$ \\
    $x \leftarrow m + (I_d - 2 e e^\top)(\hat{x} - \hat{m})$ \\
}

\Return{$(x,\,\text{bool})$}
\end{algorithm}

\subsection{Training the Drafter}
\label{training}
During training, the target DiT remains frozen as a teacher providing both noise and feature supervision, while only the drafter is updated to align its predictions and maintain temporal smoothness along the denoising trajectory.

Given a clean image $x_0$, we sample a noisy latent $x_t$ by adding Gaussian noise following the standard diffusion forward process, and then perform one denoising step using the frozen target DiT to obtain $x_{t-1}$ and its top-block feature $f_{t-1}$.
From this state, the frozen DiT produces the next-step teacher signals $(\epsilon_{\text{ref}}, f_{\text{ref}})$ (\eg the noise prediction and the corresponding top-block feature), while the drafter predicts  $\epsilon_{\text{pred}}$ and feature $f_{\text{pred}}$ for the same step conditioned on $(x_{t-1}, f_{t-1})$. 
To train the drafter, we supervise its next-step predictions with three complementary objectives.
(1) Mean squared error (MSE) on noise:
\begin{equation}
  \mathcal{L}_{\text{noise}} = \|\epsilon_{\text{pred}} - \epsilon_{\text{ref}}\|_2^2,
  \label{eq:mse}
\end{equation}
which encourages the drafter to follow the denoising dynamics of the DiT.
We align the noise prediction rather than the sampling mean, since the DiT is parameterized in noise prediction space and the mean is a deterministic function of the predicted noise and the diffusion schedule.
(2) Cosine similarity on top-block features:
\begin{equation}
  \mathcal{L}_{\text{feat}} = \frac{1}{N}\sum_{i=1}^{N}\left(1 - \frac{\langle f_{\text{pred}}^{(i)}, f_{\text{ref}}^{(i)} \rangle}{\|f_{\text{pred}}^{(i)}\|_2 \, \|f_{\text{ref}}^{(i)}\|_2 + \varepsilon}\right),
  \label{eq:cos}
\end{equation}
which enforces directional consistency between predicted and reference features, helping preserve the smooth temporal evolution observed across denoising steps.
(3) Smooth-L1 regularization on feature residuals:
\begin{equation}
  \mathcal{L}_{\text{smooth}} = \frac{1}{ND}\sum_{i=1}^{N}\sum_{j=1}^{D} \phi_\beta\!\left(f_{\text{pred}}^{(i)}[j] - f_{\text{ref}}^{(i)}[j]\right),
  \label{eq:smooth}
\end{equation}

\begin{equation}
  \phi_\beta(u)=
  \begin{cases}
    \frac{1}{2}\frac{u^2}{\beta}, & |u| < \beta,\\[6pt]
    |u| - \frac{1}{2}\beta, & \text{otherwise},
  \end{cases}
  \label{eq:huber}
\end{equation}
where $N$ and $D$ denote the token count and feature dimension, respectively, and $\phi_\beta(\cdot)$ is the Smooth-L1 function with transition parameter $\beta$. This term limits excessive feature shifts while retaining semantic variability.
Then total loss:
\begin{equation}
\mathcal{L}_{\textsc{FREE}} = \mathcal{L}_{\text{noise}} + \lambda_f \mathcal{L}_{\text{feat}} + \lambda_s \mathcal{L}_{\text{smooth}},
\label{eq:free_loss}
\end{equation}
where $\lambda_f$ and $\lambda_s$ are weighting coefficients that balance feature consistency and temporal smoothness. Although supervision is applied one step at a time, the learned transitions generalize to autoregressive rollout during inference, yielding stable multi-step forecasting.
Additional implementation details are provided in Appendix~\ref{sup_training}.

\subsection{Parallel Reflection-Based Verification}
\label{verify}
\begin{figure}[t]
  \centering
  \includegraphics[width=\columnwidth]{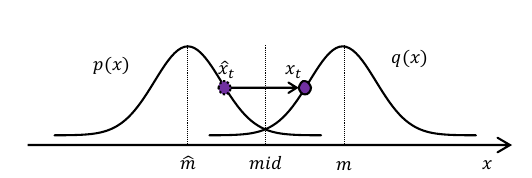}
  \caption{
   Reflection Maximal Coupling in 1D. $p(x)$ and $q(x)$ denote the draft and target distributions, respectively.
  }
  \label{fig:1d-ref}
\end{figure}

As described in~\cref{drafting}, starting from $x_a$ the drafter proposes a speculative sequence defined in~\cref{eq:drafts}.
To evaluate these proposals efficiently, the target model is run in parallel on $(x_a, \hat{x}_{a+1},\dots,\hat{x}_{b-1})$ to obtain the corresponding reference means $(m_{a+1},\dots,m_b)$ in a single forward pass.
The drafted transitions are then reconciled with the target transitions using \textit{reflection maximal coupling} (RMC)~\cite{reflection}, as outlined in~\cref{alg:verification}.
Here, the target transition is $q(x)=\mathcal{N}(m,\sigma^2 I_d)$ and the drafter proposes $p(x)=\mathcal{N}(\hat{m},\sigma^2 I_d)$. A drafted sample $\hat{x}$ is accepted with probability $\min(1, q(\hat{x})/p(\hat{x}))$.
If rejected, the sample is reflected across the hyperplane orthogonal to the mean-discrepancy direction:
\begin{equation}
  e = (\hat{m} - m)/(\|\hat{m} - m\|),
  \label{eq:direction}
\end{equation}
yielding a corrected sample whose distribution exactly matches $q(x)$.
After rejection, the remaining speculative steps are discarded and sampling resumes from the corrected state.
This correction remains unbiased as long as the drafter and the target share the same variance schedule $\{\sigma_t^2\}_{t=1}^{T+1}$, thereby preserving the exact target sampling distribution even when multiple future steps are drafted.
For intuition, in the one-dimensional case under the same variance (see~ \cref{fig:1d-ref}), the reflection operation simplifies to $x = 2\,\mathrm{mid} - \hat{x}$ with $\mathrm{mid} = \tfrac{1}{2}(m + \hat{m})$, which mirrors the rejected proposal about the midpoint between the two means.
A full correctness proof of RMC is deferred to Appendix~\ref{sup_rmc}.

\begin{figure}[t]
  \centering
  \includegraphics[width=\columnwidth]{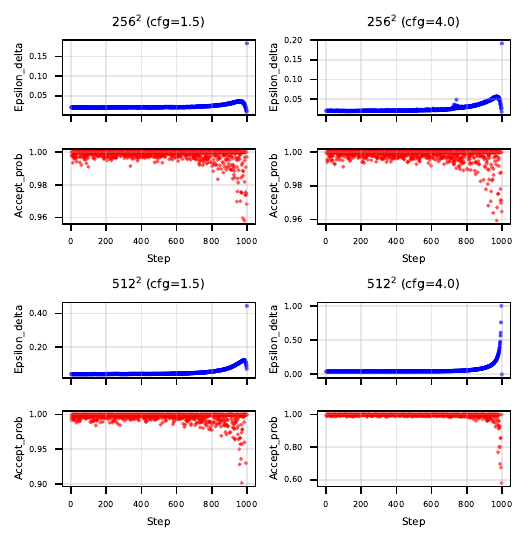}
  \caption{
   Prediction uncertainty of DiT under different resolutions and classifier-free guidance (cfg) scales.
   We run the DiT as both drafter and verifier in speculative sampling.
   Each subplot shows the mean-squared prediction deviation between the drafts and their reference predicted outputs ($\textcolor{blue}{epsilon\_delta}$), along with the acceptance probability ($\textcolor{red}{accept\_prob}$) across denoising steps.
  }
  \label{fig:acc_cha}
\end{figure}

\subsection{Uncertainty-Guided Verification Relaxation}
\label{uncertainty}
While reflection-based verification remains unbiased, the acceptance rate decreases in later denoising stages due to the increasing prediction variance of DiT, potentially limiting overall speedup.
To characterize the uncertainty dynamics in DiT’s denoising process, we perform an \textit{offline self-speculative run}, where a single DiT model both generates proposals and verifies them under the RMC rule.
During verification, we track the noise prediction deviation $\epsilon_{\Delta}(t)=\|\epsilon_{\text{pred, t}} - \epsilon_{\text{ref, t}} \|_2$, along with the acceptance probability.
As shown in Figure~\cref{fig:acc_cha}, the prediction deviation consistently increases over steps, accompanied by a corresponding drop in acceptance probability across settings.

This trend aligns with theoretical analyses showing that high signal-to-noise ratio (SNR) stages yield weaker gradient signal and limited learning capacity~\cite{vdm, karras2022elucidating}.
Consequently, noise prediction becomes inherently less precise in later steps, and small feature discrepancies are amplified through transformer layers, leading to higher uncertainty and instability during multi-step forecasting, especially at higher resolution or fidelity.
Meanwhile, at low-noise stages, the latent lies near the data manifold, making small prediction errors negligible and unlikely to accumulate over subsequent steps.
Together, these observations imply that enforcing strict equality between the drafter’s and target’s transition means $(\hat{m}, m)$ at all steps is unnecessary:
coupling should remain strict in earlier stages where deviations can propagate, and gradually relax later, where discrepancies are both more frequent and less consequential.

To realize this stage-dependent relaxation, we fit a smooth, monotonically increasing curve to the sequence $\{\epsilon_{\Delta}(t)\}_{t=1}^T$, and take its inverse (rescaled to $[0,1]$) to obtain a relaxation profile ${\lambda_t}$.
This profile compensates for the natural rise in uncertainty, maintaining roughly uniform discrepancy throughout sampling.
Notably, ${\lambda_t}$ depends primarily on image resolution and guidance scale, and remains consistent across class labels, as the uncertainty dynamics of DiTs are governed by the diffusion schedule rather than instance content.
To connect with the verification rule, recall that the acceptance probability in~\cref{alg:verification} can be expressed using the Gaussian likelihood ratio:
\begin{equation}
  \frac{\mathcal{N}(\hat{x};\,m,\,\sigma^2 I_d)}{\mathcal{N}(\hat{x};\,\hat{m},\,\sigma^2 I_d)}=\exp\!\left(\frac{(m-\hat m)^{\!\top}(\hat x-\mathrm{mid})}{\sigma^2}\right),
  \label{eq:qp}
\end{equation}
where $\mathrm{mid} = \tfrac{1}{2}(m + \hat{m})$. We scale the coupling term by $\lambda_t$, defining $v_t = \lambda_t(m_t - \hat{m}_t)$. Substituting this into the verification rule yields:
\begin{equation}
  \mathrm{acc}_\lambda = \min\!\left(1,\,\exp\!\left(\frac{v^{\!\top}(\hat{x}-\mathrm{mid})}{\sigma^2}\right)\right).
    \label{eq:acc_relax}
\end{equation}
When $\lambda_t=1$, it reduces to standard reflection coupling; smaller $\lambda_t$ values make verification more permissive.
Unlike SpeCa~\cite{SpeCa}, which adopts a loose-to-strict schedule, FREE (relax) follows the opposite strict-to-loose strategy aligned with the DiT's uncertainty dynamics, effectively balancing efficiency and fidelity, providing notable acceleration while maintaining perceptual quality comparable to standard DiT sampling.


%% file: sec/4_experiments.tex
\section{Experiments}
\label{sec:expe}

We evaluate FREE and FREE (relax) on ImageNet-$256^2$ and $512^2$ image generation using DiT-XL/2 as the base model, following the official configuration with classifier-free guidance (cfg=1.5).
The frozen DiT-XL/2 with 1000 denoising steps serves as our baseline.
Following~\cref{verify}, we fix the reverse-process variance to the posterior value $\tilde{\beta}_t I$ (as in DDPM) instead of the learned variance $\sigma_{\theta}(x_t, t)$ used in the original DiT implementation, in order to obtain a tractable formulation of reflection coupling.

We denote the number of drafts generated per speculation round in both FREE and FREE (relax) as the \emph{speculation length}.
All generation quality metrics (FID, FIDs, Precision, and Recall) are reported at a fixed speculation length of 8.
To evaluate acceleration, we report two metrics:
(1) \emph{Parallel efficiency}, defined as the ratio of total denoising steps (\eg 1000) to the actual number of DiT invocations.
(2) \emph{Wall-clock speedup}, measured as the relative reduction in per-image runtime compared to the baseline.

\begin{table}[t]
\centering
\setlength{\tabcolsep}{5pt}
\renewcommand{\arraystretch}{1.15}
\caption{
Parallel efficiency under different drafting methods and speculation lengths. ``*'' indicates no further improvement.
}
\label{tab:parallel_efficiency}
\vspace{4pt}

\subfloat[ImageNet-$256^2$]{%
\begin{tabular}{lccccc}
\toprule
Method & 2 & 4 & 8 & 16 & 32 \\
\midrule
ASD~\cite{ASD} & 1.24 & 1.35 & * & * & * \\
Frozen~\cite{reflection} & 1.28 & 1.41 & * & * & * \\
\textsc{Free} & 1.43 & 2.05 & 2.68 & 3.10 & * \\
\textsc{Free} (relax) & 1.46 & 2.22 & 3.25 & 4.01 & 4.26 \\
\bottomrule
\end{tabular}
}

\vspace{6pt}

\subfloat[ImageNet-$512^2$]{%
\begin{tabular}{lccccc}
\toprule
Method & 2 & 4 & 8 & 16 & 32 \\
\midrule
ASD~\cite{ASD} & 1.10 & 1.13 & * & * & * \\
Frozen~\cite{reflection} & 1.12 & 1.15 & * & * & * \\
\textsc{Free} & 1.39 & 1.87 & 2.29 & 2.40 & 2.46 \\
\textsc{Free} (relax) & 1.43 & 2.10 & 2.81 & 3.07 & 3.15 \\
\bottomrule
\end{tabular}
}

\end{table}

\subsection{Parallel Efficiency Analysis}
\label{parallel}
Parallel efficiency reflects the idealized theoretical acceleration bound when speculative verification can be performed in parallel at a cost comparable to one DiT forward pass, with non-neural overheads (\eg sampling and communication) ignored.
In practice, these assumptions hold only approximately under multi-GPU execution, and generating speculative drafts also introduces additional computation. As discussed in~\cref{drafting}, the drafting stage is lightweight but not cost-free.

To contextualize the efficiency trends, we also re-implement Autospeculative Decoding~\cite{ASD} and the frozen target drafting scheme of~\cite{reflection} in the same DiT-XL/2 pipeline and sampling configuration , referred to as \emph{ASD} and \emph{Frozen}, respectively.
For each resolution ($256^2$ and $512^2$), we randomly sample and fix a set of 8 ImageNet classes.
For FREE (relax), we primarily analyze speculation lengths up to 8 to match the quality evaluation setting, while results at larger lengths are also included in~\cref{tab:parallel_efficiency} for completeness.

As shown in~\cref{tab:parallel_efficiency}, ASD and Frozen exhibit only small gains from speculation length 2 to 4 and saturate thereafter, indicating a limited draft acceptance rate.
In contrast, FREE and FREE (relax) continue to benefit from increasing speculation length, demonstrating that the learned drafter provides sufficiently accurate intermediate predictions to support deeper speculative execution.
However, under reflection coupling, later drafts can be verified only if all earlier ones are accepted, causing acceptance probability to decay approximately exponentially with depth and leading to diminishing marginal gains.

On ImageNet-$256^2$, FREE reaches a parallel efficiency of $3.10 \times$ at a speculation length of 16, after which the improvement becomes marginal. 
On ImageNet-$512^2$, the efficiency reaches $2.45 \times$ at length 32, where the saturation point appears slightly later. This behavior is consistent with higher-resolution representations preserving more structured visual information, which supports valid predictions over more consecutive speculative steps and delays the efficiency plateau.
The relaxed variant FREE (relax) further improves the efficiency, reaching $3.25 \times$ on $256^2$ and $2.81 \times$ on $512^2$ at length 8. It continues to improve without a clear plateau within the tested range, benefiting from its looser verification criterion, which retains more speculative drafts before reflection correction is triggered.

\begin{figure}[t]
  \centering
   \includegraphics[width=\linewidth]{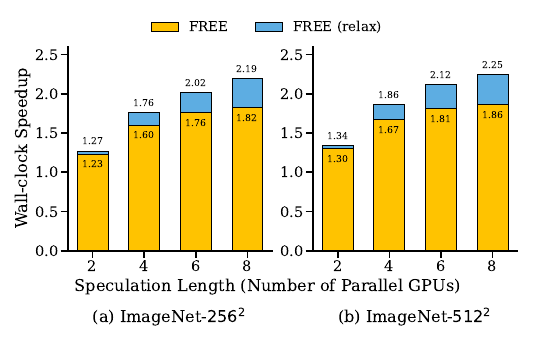}
   \caption{
   Wall-clock speedup of FREE and FREE (relax) under different speculation lengths (each corresponding to the number of parallel GPUs used during inference).
   }
   \label{fig:wall_clock}
\end{figure}

\subsection{Wall-clock Speedup}
\label{wall_clock}
For wall-clock evaluation, we randomly sample and fix 23 ImageNet classes, discard the first 3 warm-up runs, and report the average runtime over the remaining 20 runs.
The results are shown in~\cref{fig:wall_clock}. On ImageNet-$256^2$, FREE yields a $1.82 \times$ speedup over standard DiT sampling, and FREE (relax) further improves to $2.19 \times$.
On ImageNet-$512^2$, the corresponding speedups are $1.86 \times$ and $2.25 \times$, respectively.
These real-time speedups exceed the theoretical upper bound on the acceleration achievable by ASD and Frozen, as shown in~\cref{tab:parallel_efficiency}.
This trend aligns with the parallel efficiency analysis: both methods provide substantial acceleration, while FREE (relax) achieves larger gains due to its higher draft retention rate.
The gap between theoretical parallel efficiency and measured wall-clock speedup reflects the practical overhead discussed in~\cref{parallel}.
Notably, although parallel efficiency is higher at $256^2$, the wall-clock speedup is slightly larger at $512^2$.
As analyzed in~\cref{drafting}, at higher resolutions self-attention dominates the total computation, reducing the drafter’s relative overhead compared to the full DiT.
As a result, the wall-clock speedup at $512^2$ more closely approaches the theoretical efficiency.

\begin{table}[t]
\centering
\small
\caption{
Quantitative comparison on ImageNet-$256^2$ and ImageNet-$512^2$ (cfg = 1.5, speculation length = 8).
Lower FID/sFID and higher Precision/Recall indicate better performance.
}
\label{tab:generation_quality}
\vspace{4pt}

\subfloat[ImageNet-$256^2$, 50K samples]{%
\begin{tabular}{lcccc}
\toprule
Method & FID $\downarrow$ & sFID $\downarrow$ & Precision $\uparrow$ & Recall $\uparrow$ \\
\midrule
DiT & 5.372 & 11.075 & \textbf{0.753} & 0.512 \\
\textsc{Free} & 5.006 & 10.227 & \textbf{0.753} & 0.541 \\
\textsc{Free} (relax) & \textbf{4.779} & \textbf{9.587} & 0.748 & \textbf{0.566} \\
\bottomrule
\end{tabular}
}

\vspace{6pt}

\subfloat[ImageNet-$512^2$, 5K samples]{%
\begin{tabular}{lcccc}
\toprule
Method & FID $\downarrow$ & sFID $\downarrow$ & Precision $\uparrow$ & Recall $\uparrow$ \\
\midrule
DiT & 11.172 & 40.577 & \textbf{0.758} & 0.690 \\
\textsc{Free} & 11.075 & 40.574 & 0.751 & \textbf{0.700} \\
\textsc{Free} (relax) & \textbf{11.008} & \textbf{40.485} & 0.754 & 0.691 \\
\bottomrule
\end{tabular}
}

\end{table}

\subsection{Generation Quality}
\label{quality}
We evaluate the quality and diversity of generated images using the FID, sFID, Precision, and Recall metrics. The results are summarized in~\cref{tab:generation_quality}.
For ImageNet-$256^2$ (50K samples), FREE achieves generation quality comparable to the frozen DiT baseline, consistent with the theoretical guarantee of lossless verification provided by RMC~\cite{reflection, ASD}.
The relaxed variant FREE (relax) shows no perceptible degradation and even yields a slight improvement.
For ImageNet-$512^2$, we evaluate 5K generated samples due to the high computational cost of performing the full 50K evaluation.
As shown in~\cref{tab:generation_quality}, the trends are consistent with those observed at the lower resolution, confirming that both FREE and FREE (relax) maintain similar generation fidelity at higher resolutions.

\subsection{FREE (relax) under High Uncertainty}
As analyzed in~\cref{uncertainty} and illustrated in~\cref{fig:acc_cha}, stronger guidance markedly increases prediction uncertainty for DiTs, especially at high resolutions.
On ImageNet-$512^2$ with $\text{cfg}=4.0$, the acceptance probability in the \textit{offline self-speculative run} drops to $\sim 0.6$, reflecting the increased stochasticity of later denoising steps.

To assess the necessity of uncertainty-guided relaxation, we evaluate FREE (relax) against \textsc{Free} under this high-resolution, strong-guidance condition.
We randomly sample 23 ImageNet classes (as in~\cref{wall_clock}), set the speculation length to match the number of parallel GPUs, and jointly measure parallel efficiency and wall-clock speedup.
As shown in~\cref{tab:high_fid_eff}, under the high-uncertainty setting, FREE exhibits limited improvement with increasing speculation length, because strict reflection coupling becomes difficult to maintain.
At longer horizons ($L{>}4$), the efficiency gains saturate and wall-clock acceleration slightly declines, partly due to the increased inter-GPU communication overhead.
In contrast, FREE (relax) sustains consistent efficiency improvements across all tested lengths and achieves a peak wall-clock speedup of $1.76\times$.

Meanwhile, to verify whether the fidelity conclusion for FREE (relax) in~\cref{quality} still holds under stronger guidance, we further evaluate image quality at a fixed speculation length of 8 using 5K generated samples.
As reported in~\cref{tab:high_fid_quality}, FREE (relax) achieves slightly lower FID and sFID than the baseline DiT, with nearly identical precision and recall, confirming that the proposed relaxation preserves generation fidelity in this high-uncertainty regime.

\begin{table}[t]
\centering
\setlength{\tabcolsep}{5pt}
\renewcommand{\arraystretch}{1.15}
\caption{
Performance of \textsc{Free} and \textsc{Free} (relax) in the high-uncertainty setting (ImageNet-$512^2$, cfg = 4.0) across different speculation lengths.
}
\label{tab:high_fid_eff}
\vspace{4pt}
\begin{tabular}{lcccc}
\toprule
Method & 2 & 4 & 6 & 8 \\
\midrule
\multicolumn{5}{l}{\textit{Parallel Efficiency}} \\
\textsc{Free} & 1.25 & 1.46 & 1.54 & 1.59 \\
\textsc{Free} (relax) & 1.37 & 1.84 & 2.09 & 2.20 \\
\midrule
\multicolumn{5}{l}{\textit{Wall-clock Speedup}} \\
\textsc{Free} & 1.17 & 1.30 & 1.30 & 1.28 \\
\textsc{Free} (relax) & 1.28 & 1.63 & 1.76 & 1.76 \\
\bottomrule
\end{tabular}
\end{table}

\begin{table}[t]
\centering
\small
\caption{
Generation quality of \textsc{Free} (relax) on ImageNet-$512^2$ (cfg = 4.0, speculation length = 8, 5K samples).
}
\label{tab:high_fid_quality}
\vspace{4pt}
\begin{tabular}{lcccc}
\toprule
Method & FID $\downarrow$ & sFID $\downarrow$ & Precision $\uparrow$ & Recall $\uparrow$ \\
\midrule
DiT & 32.38 & 46.00 & \textbf{0.868} & \textbf{0.445} \\
\textsc{Free} (relax) & \textbf{31.19} & \textbf{45.54} & 0.863 & \textbf{0.445} \\
\bottomrule
\end{tabular}
\end{table}


%% file: sec/5_conclusion.tex
\section{Conclusion}
\label{conclusion}

Across all evaluations we conducted, FREE demonstrates efficient feature-level speculative inference to parallelize the DiT denoising process.
Its relaxed variant, guided by uncertainty estimation, further improves acceleration while preserving visual fidelity.
The results validate both the theoretical soundness and the practical effectiveness of our framework, highlighting its potential as a scalable solution for high-fidelity, low-latency diffusion generation.

To summarize, our contributions are as follows:
\begin{itemize}
    \item We propose \textbf{FREE}, a feature-level autoregressive framework for lossless parallel acceleration of DiTs via speculative inference.
    \item We develop a \textbf{feature-alignment training strategy} that ensures temporal consistency between drafter predictions and the target DiT, improving multi-step autoregressive stability.
    \item We introduce an \textbf{uncertainty-guided relaxation mechanism} that dynamically modulates draft acceptance probabilities during denoising, achieving faster inference without perceptible quality loss.
\end{itemize}

Looking ahead, future work may explore extending FREE to other diffusion modalities, and explore adaptive scheduling strategies for further improvements in inference efficiency.

%% file: sec/X_suppl.tex
\clearpage
\appendix
\onecolumn
\setcounter{page}{1}
\maketitlesupplementary

\section{Training Details}
\label{sup_training}
We train all drafter models on ImageNet-1K using a randomly sampled 30\% subset of the training split.
Each image is center-cropped to the target resolution (\eg $256^2$ or $512^2$), normalized to [-1, 1], and encoded using a frozen SD-VAE following the standard DiT preprocessing pipeline.
At each iteration, a diffusion timestep $t \in \{1, \dots, 1000\}$ is sampled. The frozen DiT-XL/2 first processes a noisy latent $x_t$, and a second forward pass at $t - 1$ yields the teacher supervision $(\epsilon_{ref}, f_{ref})$.
The drafter receives the intermediate latent $x_{t-1}$ without normalization, matching the teacher model’s inference behavior and allowing it to exploit fine-grained information along the denoising trajectory.
Classifier-free guidance also follows the DiT formulation, where 10\% of labels are randomly dropped before the teacher's first forward pass. The resulting label embedding computed by DiT is then provided directly to the drafter, which does not maintain its own label-embedding module.

Although one might consider precomputing latents and intermediate DiT features to avoid online VAE encoding, this is highly impractical.
Even at $256^2$ resolution, the top-block hidden representation of DiT-XL/2 used during training produces a flattened feature tensor of shape [1024, 1152], amounting to $\sim$ 4.7MB per sample in FP32---over an order of magnitude larger than a raw ImageNet image (typically 100–200 KB).
Moreover, because these intermediate features depend on the sampled timestep and the (possibly dropped) class label, each stored item has essentially no reuse value.
For these reasons, we generate both latents and teacher features on the fly from raw images.

The drafter itself is lightweight, containing $\sim$ 32M trainable parameters compared with the 676M parameters of the DiT-XL/2 teacher.
For the $256^2$ configuration, we train for 120 epochs and reach convergence in approximately 56.8 hours on a 4 $\times$ RTX 4090 (24GB) setup.
For $512^2$, training runs for 60 epochs and requires about 138 hours under the same hardware.
These runtimes indicate that FREE can be trained efficiently even under modest computational budgets.
Throughout training we follow the same optimization settings used for DiT.
AdamW is employed with a fixed learning rate of $1 \times 10^{-4}$, and an EMA decay of 0.9999 is applied. These standard choices provide stable optimization and help maintain close alignment between the drafter and the frozen teacher model.

\begin{figure}[h]
  \centering
  \includegraphics[width=\columnwidth]{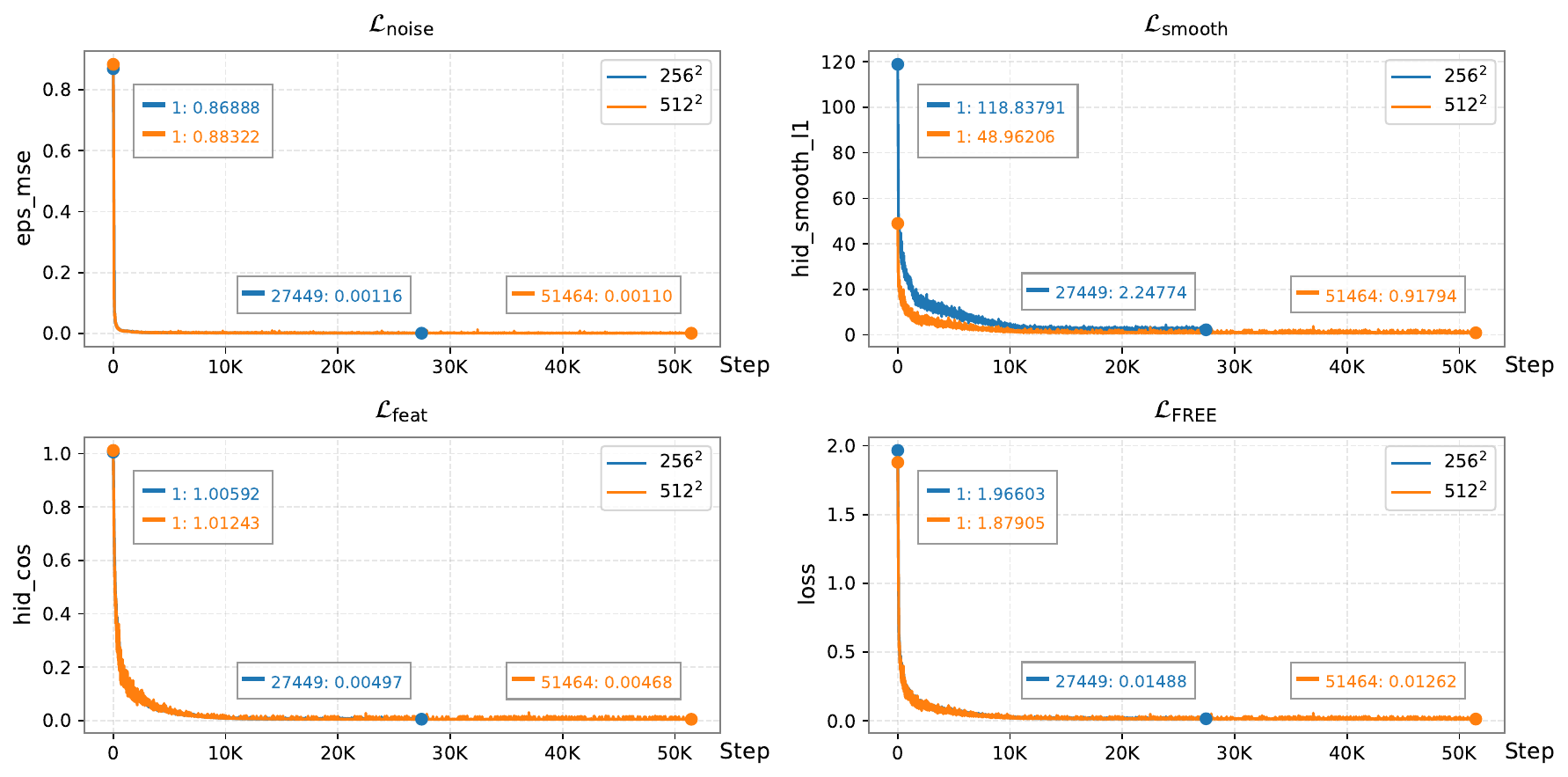}
  \caption{
  Training loss curves for the first 30 epochs at $256^2$ resolution and the first 15 epochs at $512^2$ resolution.
  }
  \label{fig:loss}
\end{figure}

As mentioned in~\ref{training}, the training loss is defined as:
$\mathcal{L}_{\textsc{FREE}} = \mathcal{L}_{\text{noise}} + \lambda_f \mathcal{L}_{\text{feat}} + \lambda_s \mathcal{L}_{\text{smooth}}$.
For the 256 resolution, we set $\lambda_f = 0.5$ and $\lambda_s = 0.005$, whereas for the 512 resolution, we kept $\lambda_f$ unchanged and increased $\lambda_s$ to 0.01.
We plotted the partial loss curves for both settings, as shown below~\ref{fig:loss}.

Meanwhile, we evaluated the model checkpoints at different training stages, and the resulting parallel efficiency is shown below~\ref{tab:scaling}. The results indicate that a larger $\mathcal{L}_{\text{smooth}}$ leads to slower convergence for the $256^2$ drafter.
\begin{table}[h]
\centering
\caption{Scaling behavior of parallel efficiency during drafter training (FREE, speculation length = 8, cfg = 1.5).}
\label{tab:scaling}

\begin{subtable}{0.48\linewidth}
\centering
\caption{$256^2$ resolution}
\begin{tabular}{c|ccccc}
\toprule
Epoch & 30 & 60 & 90 & 119 \\
\midrule
Parallel efficiency & 1.36 & 2.41 & 2.59 & 2.66 \\
\bottomrule
\end{tabular}
\end{subtable}
\hfill
\begin{subtable}{0.48\linewidth}
\centering
\caption{$512^2$ resolution}
\begin{tabular}{c|ccccc}
\toprule
Epoch & 15 & 30 & 45 & 60 \\
\midrule
Parallel efficiency & 1.88 & 2.16 & 2.29 & 2.27 \\
\bottomrule
\end{tabular}
\end{subtable}

\end{table}

\section{Theoretical Proofs for Reflection Coupling}
\label{sup_rmc}
\begin{tcolorbox}[theoremstyle]
\textbf{Theorem 1 (Maximal coupling).}
For any two probability densities $p$ and $q$ on $\mathbb{R}^d$, the total variation distance satisfies
\begin{equation}
    \|p - q\|_{TV}
    = 1 - \int \min\{p(x), q(x)\}\, dx
    = \int \max\{p(x) - q(x), 0\}\, dx.
\end{equation}
Moreover, for any coupling $(X,Y)$ with marginals $p$ and $q$,
\begin{equation}
    \mathbb{P}(X \neq Y) \;\ge\; \|p - q\|_{TV},
\end{equation}
and the equality is achievable; any coupling attaining equality is called a \emph{maximal coupling}.
\end{tcolorbox}

Recall Algorithm~\ref{alg:verification}, where the draft distribution and the target distribution are given by the two Gaussian densities:
\begin{equation}
    p(x)=\mathcal{N}(x;\hat m,\sigma^2 I_d),
    \qquad
    q(x)=\mathcal{N}(x;m,\sigma^2 I_d),
\end{equation}
which share the same covariance matrix.
In each RMC step, a draft sample $\hat{x} \sim p(x)$ is accepted ($x=\hat{x}$) with probability
\begin{equation}
    a(\hat{x}) = \min\!\left(1, \frac{q(\hat{x})}{p(\hat{x})}\right).
\label{eq:acc_prob}
\end{equation}
Otherwise, upon rejection, we apply the reflection update
\begin{equation}
    x = m + (I_d - 2ee^\top)(\hat x - \hat m),
    \qquad
    e = \frac{\hat m - m}{\|\hat m - m\|},
\end{equation}
where $e$ is referred to as the mean-discrepancy direction.
Let $\mu$ denote the distribution of the output $x$. We will show
(1) unbiasedness: $\mu = q$, and
(2) maximal coupling: $\mathbb{P}(x \neq \hat{x}) = \|p - q\|_{TV}$.

\begin{tcolorbox}[theoremstyle]
\textbf{Lemma 1 (Geometry of the reflection).}
$S(z)$ is the mirror image of $z$ across the decision hyperplane
$\{x:p(x)=q(x)\}$.
\end{tcolorbox}

\textit{Proof.}
The condition $p(x)=q(x)$ is equivalent to
\begin{equation}
    \|x-\hat m\|^2 = \|x-m\|^2
    \;\Longleftrightarrow\;
    (x - \frac{\hat{m}+m}{2})^\top \cdot \frac{\hat{m}-m}{\|\hat{m} - m\|} = 0.
\end{equation}
Hence the set $\{x:p(x)=q(x)\}$ is exactly the hyperplane
\begin{equation}
    H = \left\{x:\;
    \left\langle x - mid,\, e \right\rangle = 0
    \right\},
    \qquad mid=\tfrac{m+\hat m}{2},
\end{equation}
which passes through $mid$ and has unit normal $e$.
Define the reflection matrix
\begin{equation}
    R = I_d - 2ee^\top.
\end{equation}
It  satisfies
\begin{equation}
    R^\top R = I_d,\qquad
    Re = -e,\qquad
    Rv = v\;\; \text{for all } v\perp e.
\end{equation}
Thus $R$ is the Householder reflection across $H$.
The reflection map
\begin{equation}
    S(z) = m + R(z - \hat m)
\end{equation}
shifts $z$ to the coordinate frame centered at $\hat m$, applies the 
reflection $R$, and then shifts the result to the frame centered at $m$. 
Therefore, $S(z)$ is exactly the mirror image of $z$ across $H$.

\begin{tcolorbox}[theoremstyle]
\textbf{Lemma 2 (Density swap).}
For all $z \in \mathbb{R}^d$, $p(z) = q(S(z))$, $q(z) = p(S(z))$.
\end{tcolorbox}

\textit{Proof.}
By Lemma 1, $R$ is an orthogonal reflection, hence
\begin{equation}
    S(z) - m = R(z - \hat m) \quad\Rightarrow\quad 
    \|S(z) - m\| = \|z - \hat m\|,
\end{equation}
and similarly
\begin{equation}
    S(z) - \hat m
    = (S(z) - m) - (\hat m - m)
    = R(z-\hat m) + R(\hat m - m)
    = R(z-m)
    \;\Rightarrow\;
    \|S(z)-\hat m\| = \|z-m\|.
\end{equation}
Thus $S$ is an isometry.
Using the norm equalities established above,
\begin{equation}
\begin{aligned}
    q(S(z))
    &\propto \exp \left(-\frac{|S(z)-m|^2}{2\sigma^2}\right)
    = \exp \left(-\frac{|z-\hat{m}|^2}{2\sigma^2}\right)
    \propto p(z), \\
    p(S(z))
    &\propto \exp \left(-\frac{|S(z)-\hat{m}|^2}{2\sigma^2}\right)
    = \exp \left(-\frac{|z-m|^2}{2\sigma^2}\right)
    \propto q(z).
\end{aligned}
\end{equation}
Since the normalization constants of the two Gaussians match, we obtain the exact equalities
\begin{equation}
    p(z) = q(S(z)), \qquad
    q(z) = p(S(z)).
\end{equation}

\textbf{Unbiasedness.}
For any bounded measurable test function $f$,
\begin{equation}
    \int f(x) d\mu(x)
    = \mathbb{E}[f(x)]
    = \mathbb{E}[f(\hat{x})\mathbf{1}_{\text{acc}}]
    + \mathbb{E}[f(S(\hat{x}))\mathbf{1}_{\text{rej}}].
\end{equation}
Conditioned on $\hat{x}$, the acceptance probability is $a(\hat{x})$ and the rejection probability is $1 - a(\hat{x})$, so
\begin{equation}
    \int f(x) d\mu(x)
    = \int f(\hat{x}) a(\hat{x}) p(\hat{x})d\hat{x} + \int f(S(\hat{x}))(1 - a(\hat{x}))p(\hat{x})d\hat{x}.
\end{equation}
From~\cref{eq:acc_prob}, the acceptance part equals $\int f(x)\min(p(x),q(x)) dx$, and the rejection part is
\begin{equation}
    \int f(S(\hat{x})) (1 - a(\hat{x})) p(\hat{x}) d\hat{x}
    = \int f(S(\hat{x})) \max\!\big(p(\hat{x})-q(\hat{x}),0\big)\, d\hat{x}.
\end{equation}
Let $x = S(\hat{x})$. Since $S$ is an isometry, it is volume-preserving, hence $d\hat{x} = dx$.  
By Lemma~2, we also have $p(\hat{x}) = q(x)$ and $q(\hat{x}) = p(x)$. Thus the rejection term becomes $\int f(x) \max(q(x) - p(x), 0) dx$.
Combining the acceptance and rejection contributions,
\begin{equation}
    \int f(x) d\mu(x) = \int f(x) \left[\min(p(x), q(x)) + \max(q(x) - p(x), 0)\right] dx = \int f(x) q(x) dx.
\end{equation}
Since this holds for all test functions $f$, we obtain $\mu = q$. Unbiasedness is proved.

\textbf{Maximal coupling.}
By construction we have $\hat{x}\sim p$, and by the unbiasedness result above
the output satisfies $x\sim q$.
Using the acceptance rule~\cref{eq:acc_prob} together with Theorem 1, we obtain
\begin{equation}
    \mathbb{P}(x \neq \hat{x})
    = \int (1 - a(\hat{x}))\, p(\hat{x})\, d\hat{x}
    = \int \max\{p(x) - q(x), 0\}\, dx
    = \|p - q\|_{TV}.
\end{equation}
Hence, the RMC step realizes a maximal coupling between $p$ and $q$.

\section{Variance Setting}
\label{sup_variance}
Perhaps you have noticed that the image quality metrics (\eg FID) for DiT in~\cref{tab:generation_quality} are lower than those reported in the original DiT paper.
This is because we fix the reverse-process variance to the posterior value $\tilde{\beta}_t I_d$ (as in DDPM), rather than using the learned variance $\sigma_{\theta}(x_t, t)$ from the original DiT implementation, which provides marginal improvements in image fidelity.
It is important to emphasize that this does not weaken the DiT baseline: the underlying DiT model we evaluate is identical to the original one, and the difference comes solely from a variance-setting choice during sampling.

As discussed in~\cref{verify}, fixing the variance is necessary to facilitate the formulation of reflection coupling (RMC), which requires the draft and target distributions to share the same variance at each timestep.
To better clarify what “fixing the variance” means in the sampling procedure, we recall the standard one-step DDPM update expressed in terms of the noise-prediction model~$\epsilon_{\theta}$:
\begin{equation}
    x_{t-1} = \frac{1}{\sqrt{\alpha_t}}
    \Big(
    x_t - \frac{1-\alpha_t}{\sqrt{1-\bar\alpha_t}}\,
    \epsilon_\theta(x_t, t)
    \Big)
    + \sigma_t z,\quad z\sim \mathcal{N}(0,I).
\label{eq:ddpm_update}
\end{equation}
where $\sigma_t = \sqrt{\tilde{\beta}_t I_d}$ is precisely the posterior variance term, ensuring that the reverse dynamics remain consistent with the assumed diffusion posterior.
This formulation makes explicit that the stochasticity injected at each step is entirely governed by~$\sigma_t$.
Thus, “fixing the variance’’ simply means choosing a particular, predefined schedule for these~$\sigma_t$ values.

In fact, the posterior choice $\sigma_t^2 = \tilde{\beta}_t I_d$ is merely one particular fixed-variance schedule, not a fundamental requirement.
From the quality standpoint, more expressive fixed-variance schedules beyond this posterior choice could be adopted.
One possibility is to analyze the temporal patterns in DiT’s learned variance and design a fixed variance schedule that mimics these patterns, ensuring the fixed variance adapts consistently with the learned behavior of the DiT.
Alternatively, a small auxiliary network can be employed to predict the variance for each denoising step based on the current state and timestep, without interacting with the generative model.
These approaches would offer flexibility and could potentially improve image fidelity while preserving the tractability of reflection coupling.
Moreover, from the perspective of acceleration, the minimum posterior variance is actually the least favorable for speculative inference. Intuitively, a very small variance makes the reverse transitions too concentrated, which reduces the overlap between the draft and target kernels, leading to poorer acceptance.

Overall, the use of a fixed variance in our work is solely motivated by the analytical requirements of RMC, and should not be interpreted as a limitation of our method. More expressive and fidelity-oriented variance schedules can be incorporated seamlessly, without affecting the structure or generality of the proposed framework.

\section{Stochasticity Hyperparameter}
A complementary mechanism for enhancing speculative inference is to introduce a stochasticity hyperparameter $\epsilon$ into the reverse dynamics.
To place this idea in context, we briefly review how diffusion models are typically defined through a forward noising process that gradually transforms data into an approximately Gaussian distribution.
This forward process is described by the stochastic differential equation (SDE):
\begin{equation}
    dX_t = f_t X_t dt + g_t dW_t, \qquad t \in [0, 1],
\end{equation}
where $f_t$ and $g_t$ control signal decay and noise injection, and $W_t$ is a standard Wiener process. Let $q_t$ denote the marginal distribution of $X_t$, and satisfy
\begin{equation}
    X_0 \sim q_0 = p_{\text{data}}, \qquad X_1 \sim q_1 \approx \mathcal{N}(0, I).
\end{equation}
By the theory of time reversal for diffusion processes, the corresponding denoising process $Y_t := X_{1-t}$ evolves according to the reverse-time SDE:
\begin{equation}
    dY_t = \left(-f_{1-t} Y_t + g_{1-t}^2 s_{1-t}(Y_t) \right) dt + g_{1-t} dB_t,
\end{equation}
where $s_t(x) = \nabla_x \log q_t(x)$ is the Stein score (\ie the gradient of the log-density) of the intermediate marginal distribution, and $B_t$ is another independent Wiener process. If the true score were available, integrating this reverse SDE would recover samples from the data distribution.
Since the reverse-time SDE cannot be solved analytically, sampling relies on a time-discretized numerical approximation of the continuous reverse dynamics.
Let the interval $[0, 1]$ be partitioned into $K$ steps:
\begin{equation}
    0 = t_0 < t_1 < \dots < t_K = 1, \qquad \gamma_k = t_k - t_{k-1}.
\end{equation}
Defining the reverse-time drift
\begin{equation}
    b_t(x) := -f_{1-t} x + g_{1-t}^2 s_{1-t}(x),
\end{equation}
a single reverse step can be approximated using the Euler–Maruyama scheme, which induces a Gaussian transition kernel
\begin{equation}
    q(x_{t_{k-1}}|x_{t_k}) = \mathcal{N} (m_k(x_{t_k}), \sigma_k^2I),
\end{equation}
with drift mean and diffusion variance given by
\begin{equation}
    m_k(x_{t_k}) = x_{t_k} + \gamma_k b_{t_k}(x_{t_k}), \qquad \sigma_k^2 = \gamma_k g_{1-t_k}^2.
    \label{eq:mean_var}
\end{equation}
In practice, however, the true score $s_t(x)$ is not accessible.
Diffusion models therefore replace it with a learned approximation $s_{\theta}(x, t)$ when evaluating the drift and simulating the reverse Markov chain during sampling.

A tunable stochasticity hyperparameter $\epsilon \ge 0$ can be introduced by generalizing the reverse SDE. Following prior work, we obtain a family of reverse-time processes
\begin{equation}
    dY_t^{(\epsilon)} = b_t^{(\epsilon)}(Y_t^{(\epsilon)}) dt + \epsilon g_{1-t} dW_t, \qquad 
    b_t^{(\epsilon)}(x) := - f_{1-t} x + \frac{1+\epsilon^2}{2} g_{1-t}^2 s_{1-t}(x).
\end{equation}
Under ideal continuous-time conditions, and assuming access to the exact score $s_t(x)$, this construction guarantees that for any $\epsilon \geq 0$,
\begin{equation}
    Y_{1-t}^{(\epsilon)} \sim q_t, \qquad \forall t\in[0,1],
\end{equation}
so that all members of the family share the same set of marginal distributions.
The case $\epsilon=1$ recovers the standard reverse-time SDE; $\epsilon=0$ removes the stochastic term and yields a deterministic ordinary differential equation (ODE); $0<\epsilon<1$ interpolates between the ODE and SDE regimes; and $\epsilon>1$ corresponds to a more strongly stochastic reverse process.

After introducing the stochasticity parameter $\epsilon$, the drift mean and diffusion variance in the corresponding discrete reverse (\cref{eq:mean_var}) update becomes
\begin{equation}
    m_k(x_{t_k}) = x_{t_k} + \gamma_k b_{t_k}^{\epsilon}(x_{t_k}), \qquad \sigma_k^2 = \epsilon^2 \gamma_k g_{1-t_k}^2,
    \label{eq:mean_var_gamma}
\end{equation}
so different values of $\epsilon$ effectively modulate the level of stochasticity in the discrete reverse transitions.

In speculative sampling, the acceptance probability is determined by how well the drafter’s proposals align with the transition distribution of the target model.
The stochasticity hyperparameter $\epsilon$ provides a convenient means for adjusting this alignment.
When $\epsilon$ is too small, the reverse kernels become overly sharp and the overlap between the drafter and target transitions decreases.
Conversely, excessively large $\epsilon$ injects too much noise, pushing proposals into low-probability regions of the target model, where even small mean mismatch leads to sharply reduced acceptance.
In practice, intermediate values of $\epsilon$ often strike a favorable balance, leading to more stable behavior and higher acceptance rates in speculative inference.

We can further quantify how the expected acceptance probability depends on the stochasticity parameter $\epsilon$.
Let the proposal (drafter) distribution and the target distribution be denoted by $p(x)=\mathcal{N}(x; m^p, \sigma^2 I)$ and $q(x)=\mathcal{N}(x; m^q, \sigma^2 I)$, respectively. Under RMC, the mean acceptance probability is
\begin{equation}
    \mathbb{E}[a] = \int p(x) \min \left(1, \frac{q(x)}{p(x)} \right) dx = 2\Phi \left( \frac{-|m^p-m^q|}{2 \sigma} \right),
\end{equation}
which is entirely determined by the quantity
\begin{equation}
    \Delta := \frac{m^p-m^q}{\sigma}.
\end{equation}
Substituting the expressions from~\cref{eq:mean_var_gamma}, we obtain
\begin{equation}
    \| \Delta \|^2
    = \left\| \frac{\gamma \frac{1+\epsilon^2}{2} g^2 (s^p(x) - s^q(x) )}{\epsilon \sqrt{\gamma} g} \right\|^2
    = \frac{1}{4} \gamma \left(\epsilon + \frac{1}{\epsilon} \right)^2 g^2 \|s^p(x) - s^q(x)\|^2,
    \label{eq:epsilon}
\end{equation}
where $s^p(x)$ and $s^q(x)$ denote the scores computed by the proposal model and the target model, respectively.
Although the analytic expression in~\cref{eq:epsilon} suggests that $\epsilon=1$ minimizes $\left(\epsilon + \frac{1}{\epsilon} \right)^2$ and is therefore optimal in the idealized continuous-time setting, this conclusion does not hold in practical diffusion models.
In discrete time, the dominant source of deviation between the proposal and target transitions arises from drift errors, which include both score-approximation error and the numerical discretization error of Euler–Maruyama.
These errors enter the drift term through the factor
\begin{equation}
    \frac{1+\epsilon^2}{2}.
\end{equation}
Consequently, when $\epsilon > 1$ they are amplified quadratically, causing the proposal mean to deviate substantially from that of the target.
When $\epsilon<1$, this amplification is suppressed and the drift becomes more stable, leading to a significantly smaller mean mismatch.
As a result, the empirical optimum in practical speculative sampling tends to shift toward values $\epsilon<1$.